\relax
\documentclass[letterpaper]{article} 
\usepackage{aaai22}  
\usepackage{times}  
\usepackage{helvet}  
\usepackage{courier}  
\usepackage[hyphens]{url}  
\usepackage{graphicx} 
\usepackage{natbib}  
\usepackage{caption} 
\DeclareCaptionStyle{ruled}{labelfont=normalfont,labelsep=colon,strut=off} 
\usepackage{algorithm}
\usepackage{algorithmic}
\usepackage{bbm}
\usepackage{amsmath}

\usepackage{newfloat}
\usepackage{listings}
\floatstyle{ruled}
\newfloat{listing}{tb}{lst}{}
\floatname{listing}{Listing}
\title{Gaze Estimation with Eye Region Segmentation and Self-Supervised\\Multistream Learning}

\author{
    Zunayed Mahmud$^{1,2}$, Paul Hungler$^{2}$, Ali Etemad$^{1,2}$
    }
\affiliations{
    $^1$Department of Electrical and Computer Engineering\\
    $^2$Ingenuity Labs Research Institute\\
    Queen's University, Kingston, Canada\\
    \{zunayed.mahmud, paul.hungler, ali.etemad\}@queensu.ca
}

\usepackage{bibentry}

\begin{document}

\maketitle

\begin{abstract}
We present a novel multistream network that learns robust eye representations for gaze estimation. 
We first create a synthetic dataset containing eye region masks detailing the visible eyeball and iris using a simulator. 
We then perform eye region segmentation with a U-Net type model which we later use to generate eye region masks for real-world eye images. Next, we pretrain an eye image encoder in the real domain with self-supervised contrastive learning to learn generalized eye representations. Finally, this pretrained eye encoder, along with two additional encoders for visible eyeball region and iris, are used in parallel in our multistream framework to extract salient features for gaze estimation from real world images. We demonstrate the performance of our method on the EYEDIAP dataset in two different evaluation settings and achieve state-of-the-art results, outperforming all the existing benchmarks on this dataset. We also conduct additional experiments to validate the robustness of our self-supervised network with respect to different amounts of labeled data used for training. 

\end{abstract}

\section{Introduction}
Eye gaze acts as a non-verbal behavioral sign that can indicate human intention, attention, and interests. In addition to helping us perceive the environment, gaze has various applications in human behaviour analysis \cite{ishii2016prediction}, human computer interaction (HCI) \cite{andrist2014conversational}, human robot interaction (HRI) \cite{moon2014meet}, cognitive science \cite{9542977}, and virtual reality \cite{konrad2020gaze}. Thus, accurate estimation of gaze has gained attention over the years, making gaze estimation a well-established area of research.

Many prior works in gaze estimation relied on lightweight machine learning algorithms \cite{lu2011head, wood2016learning}. They were mostly biased with constrained environments and fixed head-pose. Recent methods \cite{zhang2015appearance,zhang2017mpiigaze, park2018learning} have leveraged deep neural networks to perform gaze estimation and shown performance improvement. 
Despite the progress, there still exists a number of underlying open problems:
\begin{itemize}
\item \textbf{Eye Region Registration:}
The geometric shape and orientation of \textit{particular regions} of the eye can provide strong information cues for gaze estimation. Thus, accurate segmentation of eye regions such as the iris could enhance gaze estimation. Most existing gaze datasets do not provide ground truths that could be readily used for eye region segmentation.
\item \textbf{Labeled Data:} 
As is the case with any machine learning problem, obtaining labeled data for gaze estimation can be difficult, time-consuming, and expensive. This is particularly challenging for gaze estimation as eye images are mainly cropped from larger facial images, making them small and with low resolutions, which in turn makes labeling more difficult and prone to errors.
\end{itemize}

In this work, we tackle these challenges by proposing a model for self-supervised gaze estimation. 
Our proposed solution first performs eye region segmentation using a U-Net type network \cite{ronneberger2015u} exclusively trained on a synthetic dataset created using UnityEyes \cite{wood2016learning}. 
This eye segmenter learns to create eye masks detailing the visible eyeball and the iris based on the synthetic dataset, which we 
later use to generate eye masks for the \textit{real} dataset. This domain transfer is a necessary step in our pipeline as large gaze datasets with ground-truth eyeball and iris segment labels do not exist. Hence the synthetic dataset was created with these labels. We then propose a multistream gaze estimator network which we first pretrain using self-supervised contrastive learning and then fine-tune for gaze estimation. The multistream gaze estimator network takes a single eye image and corresponding eye masks (created by the eye segmenter) as input and regresses the 3D gaze vectors. 

In a nutshell, we make the following contributions:
(\textbf{1}) We are the first to propose the usage of eye region masks as a feature input for gaze estimation. To effectively train a model for eye region segmentation on real images, we generate and use a synthetic dataset, which we then use for segmentation in the \textit{real} domain.
(\textbf{2}) We introduce a novel multistream deep neural network that utilizes both appearance and geometric factors by extracting features from raw eye image and eye masks independently to perform gaze estimation. We pretrain this model in a self-supervised manner for better generalization and reduced reliance on labels. Our model outperforms the existing supervised methods on a benchmark dataset.


\section{Related Work}
\noindent \textbf{Gaze Estimation.}
Gaze estimation can be categorized into three different approaches: feature-based, model-based, and appearance-based. Feature-based methods \cite{huang2014building, xiong2014eye} generally rely on hand-crafted features such as pupil centers, eye corners, and iris edges extracted from eye images. Model-based approaches \cite{wood20163d,wood2014eyetab,park2018learning} on the other hand, aim to fit 3D eye models to eye images. Finally, appearance-based gaze estimation methods \cite{baluja1994non, tan2002appearance} take raw eye images and learn to map gaze as a 2D point or a 3D angular vector.

Recent works on gaze estimation \cite{zhang2015appearance, zhang2017mpiigaze, park2018deep} rely on Convolutional Neural Networks (CNNs) and have shown significant performance improvements. A multitask learning network was developed in \cite{yu2018deep} where eye landmark detection was performed as an auxiliary task. In \cite{krafka2016eye}, a multistream strategy was proposed where separate convolutional feature extractors were used for both eye and face images. The approach improved on the results of \cite{zhang2015appearance} by a large margin. In \cite{wood2016learning} a real-time synthetic eye image simulator, UnityEyes, was developed to generate millions of synthetic images to create scope for domain transfer learning in gaze estimation. This work shows competitive performance in cross-dataset experiment. In \cite{shrivastava2017learning} the domain gap between synthetic and real eye images was addressed and reduced by domain adversarial learning. In a subsequent work by \cite{lee2018simulated+}, the domain adaptation results were further improved by leveraging bidirectional mapping between synthetic and real domains.
 
\noindent \textbf{Contrastive Learning.}
Contrastive learning is primarily a self-supervised learning (SSL) method which aids deep neural networks to learn a better latent representation by utilizing augmented unlabeled data. A recent contrastive learning method, SimCLR \cite{chen2020simple}, demonstrates that by applying contrastive loss and maximizing the agreement between two augmented versions of the same image, it can facilitate neural networks to learn more useful visual representations. In \cite{grill2020bootstrap} two neural networks were used which are referred to as online and target networks, where the online network predicts the output representation of the target network given two augmented versions of the same image as input. Recent applications of contrastive learning include medical image segmentation \cite{chaitanya2020contrastive, zhao2021contrastive}, facial expression detection \cite{roy2021self}, pose estimation \cite{spurr2021self}, and others. Nonetheless, despite the high likelihood that contrastive learning could aid in better generalization of gaze estimators in dealing with challenging scenarios and low quality inputs, it has not yet been explored in this area.


\begin{figure*}[t]
    \centering
    \includegraphics[width=0.9\linewidth]{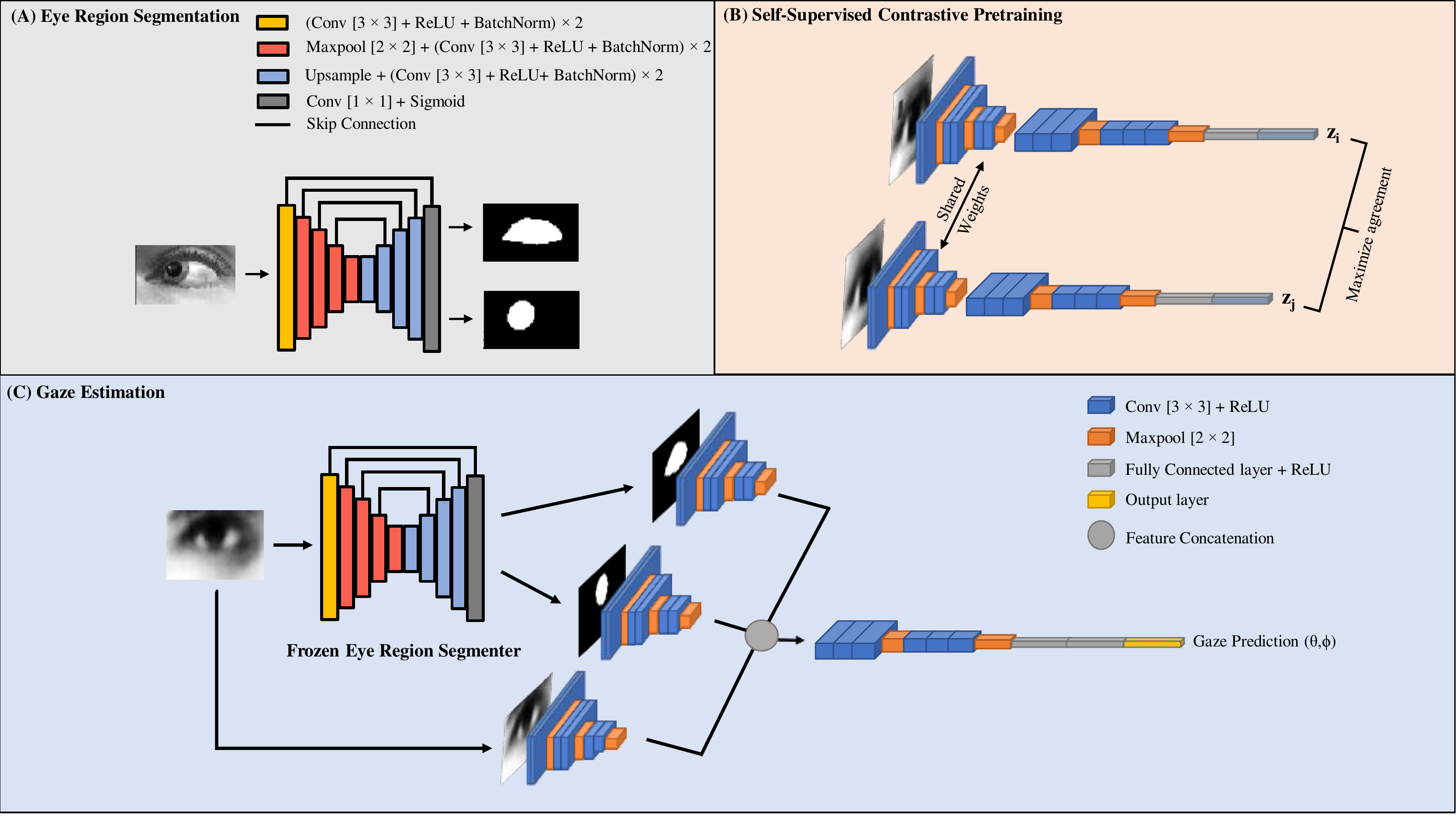} 
    \caption{The proposed framework for our method. First, we train an eye region segmenter using a synthetic database which we create using a simulator \cite{wood2016learning}. Next, we pretrain the eye image encoder by self-supervised contrastive learning. Finally, we fine-tune the multistream network for gaze estimation.}
    \label{fig:architecture}
\end{figure*}

\section{Method}
\noindent \textbf{Model Structure and Overview.}
We aim to develop a gaze estimator that can focus on salient regions of the eye, which we hypothesize contain important gaze-related information. While one approach would be to rely on eye landmarks \cite{park2018learning}, landmark detection is time-consuming and sensitive to noise. Hence, to focus the learning of the gaze estimator network without the need to detect individual and fine-grained landmarks, we rely on the iris and visible eyeball masks, in addition to the whole image of the eye. Accordingly we design a multistream pipeline to learn representations from each of these three inputs independently, which can then be combined for gaze estimation. First, an eye segmentation module is trained to detect the visible eyeball and the iris masks. As gaze datasets with ground-truth labels for eyeball and iris regions are very rare, we train the eye segmenter on an auxiliary synthetic dataset, which we discuss in detail later in Section 4. The output masks are then used to train two encoders for learning representations from the iris and visible eyeball segments as two of the three streams of our multistream pipeline. Next, we pretrain an encoder in a contrastive self-supervised manner to learn generalized representations from real input images of the whole eye. Given the binary nature of the iris and visible eyeball masks, we do not use self-supervised learning for pretraining the two mask encoders as our empirical experiments demonstrate no performance boost with such pretraining. This encoder, along with the other two encoders are used in parallel in our multistream setup to extract relevant features. The output representations from these three encoders are then fused in the feature dimension and utilized for final gaze estimation. Below we describe the details of each of the modules of our pipeline.

\noindent \textbf{Eye Region Segmentation.}
As mentioned above, we aim to adopt iris and visible eyeball masks in our multistream pipeline to provide useful gaze-related information in addition to the standard way of learning gaze from the entire eye image. To this end, a network is required to segment eye images into the two desired masks. We thus adopt a U-Net type model \cite{ronneberger2015u} with two output channels, one for the iris mask and the other for the visible eyeball mask. The architecture details of this model are presented in Fig.~\ref{fig:architecture}(A). To train this network, we use UnityEyes simulator \cite{wood2016learning} and generate a synthetic image dataset given the scarcity of detailed and accurate datasets with labeled iris and visible eyeball regions. The network is trained from scratch using MSE loss. 

\begin{table}
\small
\centering
\caption{Description of transformations.}
\label{table:3}

\begin{tabular}{l c }
 \hline
 \textbf{Transformations} & \textbf{Parameters} \\ 
 \hline\hline
 Gaussian noise & $\sigma = (0 - 10.0)$, $\mu = 0$ \\ 
 Gaussian blur & $\sigma = (0 - 2.0)$, filter size = $(3 \times 3)$ \\
 Cutout & $h,w = (0 - 10.0)~\text{px}$ \\
 Downscale & $(1x - 2x)$ \\
 Random lines & $(0 - 2)$ \\
Contrast change & - \\
 \hline

\end{tabular}

\end{table}

\noindent \textbf{Self-Supervised Learning of Eye Images.}
To utilize a model for effective learning of real eye images, we rely on self-supervised pretraining. This design choice is made to enable the encoder to be able to learn strong representations in the presence of different variations and challenging conditions (low resolution, noise, occlusions, etc.) often available in eye images which are generally cropped from facial images. To this end, we first define a set of transformations as described in Table~\ref{table:3} which are randomly applied to augment the images. Following \cite{chen2020simple}, we only draw positive pairs within a minibatch $N$ and the other 2$(N-1)$ samples are considered to be negatives. To preserve the gaze direction, we avoid applying any geometric transformation which might potentially harm the eye shape and orientation. Next, as shown in Fig.~\ref{fig:architecture}(B), a contrastive approach is used to learn useful representations using the positive and negative augmented samples. The contrastive loss \cite{chen2020simple} is used for pretraining the eye encoder, which is defined as:
\begin{equation}
    L_{i,j} = -log \frac{exp(cosine(z_{i}, z_{j})/\tau)}{\sum_{k=1}^{2N} \mathbbm{1} _{[k \neq i]} exp(cosine(z_{i}, z_{k})/\tau)}
\end{equation}
where $i$ and $j$ are the positive pairs, $\mathbbm{1}_{[k \neq i]} \in [0,1]$ is a function which results in $1$ only if $i \neq k$, and the temperature $\tau$ is a scaling parameter for the cosine similarity, set to 0.1.

\noindent\textbf{Encoder, Fusion, and Gaze Estimation Details.} The details of the encoders used for each of the streams are presented in Fig.~\ref{fig:architecture}(C). It should be emphasized that the architecture details of all three encoders are identical with their training being the only difference. Subsequent to these three encoders, the learned representations are concatenated in the feature dimension and subsequently go through a final encoding process prior to gaze estimation. The 3 fully connected (FC) layers used for gaze estimation consist of 256, 128, and 2 units respectively. Output gaze is estimated in pitch ($\theta$) and yaw ($\phi$) angles based on common practice in the field \cite{zhang2015appearance, park2018deep}. The entire pipeline, including the encoders and gaze estimator are trained using MSE loss.

\section{Experiments and Results}

\noindent \textbf{Dataset and Preprocessing Steps.}
As mentioned earlier in Section 3, to perform eye region segmentation, we generate a synthetic dataset using the eye simulator, UnityEyes \cite{wood2016learning}. The simulator provides synthetic eye images and their corresponding landmarks, which we use to generate the output masks. We generate a total of 60,000 synthetic images. To bring the domain of these synthetic data closer to that of real images, we apply the same augmentations presented earlier in Table~\ref{table:3}. It should be re-emphasized, however, that this is merely carried to as a preprocessing step, and contrastive learning is \textit{not} used for training the U-Net with the synthetic data.

For training the eye image encoder as well as gaze estimation, we use the EYEDIAP dataset \cite{funes2014eyediap}.
There are multiple sessions in the dataset with different illumination conditions, continuous gaze targets, and head pose. The low resolution images and extreme occlusions make this dataset highly challenging for gaze estimation task. We evaluate our proposed solution on the screen target gaze session (CS/DS) and static and mobile (SP/MP) head-pose scenario. The input eye images are taken from the VGA video provided for this session. The raw images are first normalized to have frontal head-pose and then converted to a fixed size of 36$\times$60. Later, these images are converted to gray-scale and histogram equalized. We follow the same person-independent or leave-one-subject-out (LOSO) protocol as in \cite{funes2016gaze}, as well as 5-fold cross-validation following \cite{park2018deep}. 

\begin{table}[t]
\small
\centering
\caption{Performance comparison on EYEDIAP dataset.}
\label{table:1}

\begin{tabular}{l c c }
 \hline
 \textbf{Methods} & \textbf{LOSO} & \textbf{5-fold} \\ [0.5ex] 
 \hline\hline
 \cite{zhang2015appearance} & 7.60 & -\\ 
 \cite{zhang2017mpiigaze} & 6.3 & - \\
\cite{yu2018deep} & 6.50 & - \\
  \cite{park2018deep} & - & 10.3 \\
 \cite{wang2019generalizing} & - & 9.90 \\
 \textbf{Ours (Multistream Network)} & 6.29 & 6.48 \\
 \textbf{Ours (SSL + Multistream Network)} & \textbf{6.15} & \textbf{6.34} \\  [1ex] 
 \hline

\end{tabular}

\end{table}


\noindent \textbf{Implementation.}
We train the eye segmenter U-Net, the 3 encoders, and the gaze estimator network using Adam optimizer with an initial learning rate of 0.00001. For segmentation, we train the network for 50 epochs using a step learning rate decay with a step size of 5 and a factor of 0.1. For the gaze estimator network, first we pretrain the eye encoder with contrastive learning as discussed earlier, for 50 epochs. Here, we again use Adam optimizer and a learning rate of 0.0001 with cosine learning rate decay. We then train the multistream gaze estimator for 25 epochs where we keep the eye image encoder frozen for 5 epochs and then fine-tune the whole network for 20 epochs. At this step, we use a plateau learning rate decay with decay factor of 0.1 and patience of 3. We train our multistream network with the same hyperparameters to create a baseline without the self-supervision. We use a batch size of 32 for the segmentation task and 128 for the contrastive learning and gaze estimation task. Our networks are implemented using PyTorch with a single Nvidia 2080 Ti GPU.  

\noindent \textbf{Results.}
We perform both LOSO and 5-fold evaluation on the EYEDIAP dataset to evaluate the gaze estimation error achieved by our proposed self-supervised multistream network. We compare the performance of our proposed framework against the existing state-of-the-art solutions on this dataset. The results as shown in Table~\ref{table:1} are reported in mean angular errors across subjects and folds. It is observed from the table, that our method outperforms the existing benchmarks by a large margin in both LOSO and 5-fold evaluations. Specifically, our supervised multistream baseline improves the highest benchmark result by a mean angle error of 3.42$^{\circ}$ in the 5-fold evaluation while achieving similar performance as \cite{zhang2017mpiigaze} with LOSO. With contrastive learning, the proposed supervised baseline is further improved by 0.14$^{\circ}$ in both LOSO and 5-fold experiments.

\noindent \textbf{Impact of Output Labels.}
To obtain a clear understanding of the effectiveness of the contrastive learning strategy on our multistream network beyond improvement of the results, we also conduct an experiment to examine the sensitivity of our network by varying the amounts of output labels. We choose a range of labeled data to fine-tune our network and compare the performance of gaze estimation. The analysis as reported in Table~\ref{table:2} reflects the performance comparison using different amounts of output labels used in the downstream fine-tuning. It can be observed that when we use only 75\% and 50\% of the labeled data, the gaze estimation error is still quite low. As we keep decreasing the percentage of labeled data to only 25\%, our network can still maintain a very stable performance and outperform existing state-of-the-art methods on the 5-fold evaluation. This validates the robustness of our multistream network through the self-supervised step.  

\begin{table}
\small
\centering
\caption{Performance of our method using different amounts of output labels.}
\label{table:2}

\begin{tabular}{l c }
 \hline
 \textbf{Labeled Data(\%)} & \textbf{Angle Error$^{\circ}$} \\ [0.5ex] 
 \hline\hline
 100\% & 6.34 \\ 
 75\% & 6.86 \\
 50\% & 7.01 \\
 25\% & 7.45 \\
 \hline

\end{tabular}

\end{table}

\section{Conclusion and Future Work}
This paper presents a novel multistream neural network for gaze estimation. 
Our model relies on a mask of the visible eyeball, a mask of the iris, and the entire image of the eye. To obtain the eye masks, we generate a synthetic dataset and use it to train a U-Net model for eye region segmentation. This segmenter is then frozen and used to generate the masks from real eye images. The masks, as well as the original eye image, are then passed to three separate encoders. The obtained embeddings are then fused to be used for gaze estimation. To allow the eye image encoder to learn more generalized representations, we use contrastive learning for pretraining. We validate our method on the EYEDIAP dataset and demonstrate that our method outperforms other works in the area with both 5-fold and LOSO schemes, while performing consistently and robustly even when small subsets of output labels are used for downstream gaze estimation. For future work, we plan to improve the eye region segmentation and perform more in-depth analysis of the segmentation performance. We would also like to further explore the effects of self-supervision on our gaze estimation model and experiment with additional augmentations.

\bibliography{aaai22}

\end{document}